\pdfoutput=1
\documentclass[11pt]{article}
\usepackage{acl}

\usepackage{times}
\usepackage{latexsym}

\usepackage[T1]{fontenc}
\usepackage[utf8]{inputenc}
\DeclareUnicodeCharacter{266B}{\textmusicalnote}

\usepackage{microtype}
\usepackage{inconsolata}

\usepackage{amsmath}
\usepackage{amssymb, bm}
\usepackage{multirow}
\usepackage{booktabs}
\usepackage{array}
\usepackage{graphicx}
\usepackage{amsfonts}
\usepackage{hyperref}
\usepackage{xspace}
\usepackage{tabularx}
\usepackage{algorithm}
\usepackage{algpseudocode}
\usepackage{array}
\usepackage{xcolor}
\usepackage{comment}
\usepackage{booktabs}

\renewcommand*{\thefootnote}{\fnsymbol{footnote}}
\title{Probing-RAG: Self-Probing to Guide Language Models in Selective Document Retrieval}

\author{Ingeol Baek, ~Hwan Chang, ~Byeongjeong Kim, ~Jimin Lee, ~Hwanhee Lee\textsuperscript{$\dagger$}\\
{Department of Artificial Intelligence, Chung-Ang University, Seoul, Korea} \\
\texttt{\{ingeolbaek, hwanchang, michael97k, ljm1690, hwanheelee\}@cau.ac.kr} \\
}

\begin{document}
\maketitle
\footnotetext{\textsuperscript{$\dagger$}Corresponding author.}
\renewcommand*{\thefootnote}{\arabic{footnote}}
\begin{abstract}
Retrieval-Augmented Generation (RAG) enhances language models by retrieving and incorporating relevant external knowledge. However, traditional retrieve-and-generate processes may not be optimized for real-world scenarios, where queries might require multiple retrieval steps or none at all. In this paper, we propose a Probing-RAG, which utilizes the hidden state representations from the intermediate layers of language models to adaptively determine the necessity of additional retrievals for a given query. By employing a pre-trained prober, Probing-RAG effectively captures the model's internal cognition, enabling reliable decision-making about retrieving external documents. Experimental results across five open-domain QA datasets demonstrate that Probing-RAG outperforms previous methods while reducing the number of redundant retrieval steps. 

\end{abstract}

\section{Introduction}
\label{lab:intro}
Large Language Models (LLMs) have demonstrated remarkable performance across a variety of tasks ~\citep{brown2020language, ouyang2022training, touvron2023llama, zhao2023survey}. However, they still face challenges such as hallucinations~\citep{ji2023survey,zhang2023siren, huang2023survey} and factual errors~\citep{min-etal-2023-factscore, manakul2023selfcheckgpt, wang2023survey}. Retrieval-Augmented Generation (RAG) leverages external knowledge related to the query through information retrieval steps to mitigate these issues.

Typically, the RAG pipeline follows a retrieve-and-generate process~\cite{guu2020retrieval,izacard2020leveraging,singh2021end, izacard2023atlas, lazaridou2022internet, shi2023replug}, fetching relevant documents based on the user’s input before generating a response. 
While this approach works well for simple single-hop QA~\citep{ram-etal-2023-context}, certain problems cannot be resolved in a single step and require multiple retrieval steps~\cite{zhang2023repocoder, shao2023enhancing, cheng2024lift}.
On the other hand, some examples can be addressed using the LLM's internal parametric knowledge without requiring any retrieval steps.

\begin{figure}[!t]
\centerline{\includegraphics[width=\linewidth]{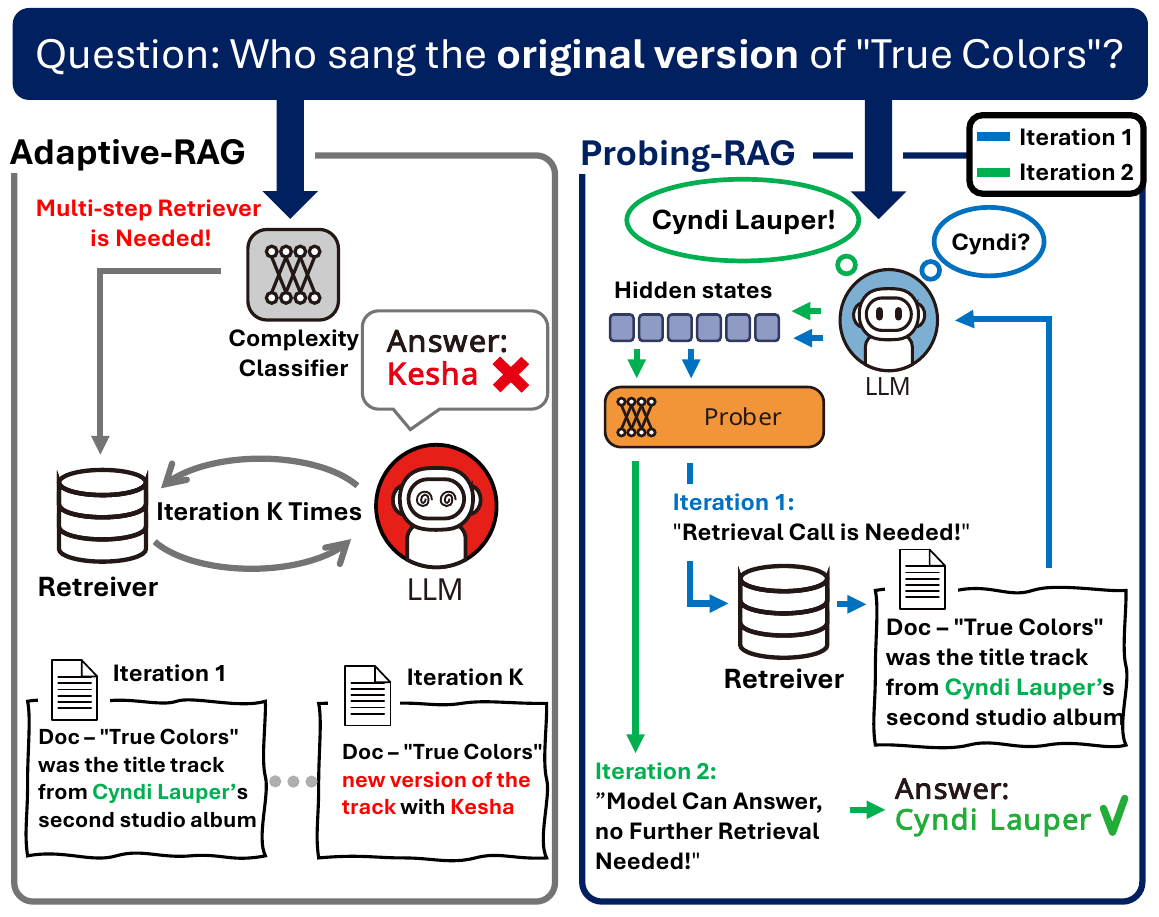}}
\caption{The left example illustrates how redundant retrieval steps, guided by an external query complexity classifier that does not reflect the LLM's internal knowledge, can lead to wrong answers. In contrast, the right example shows that the model uses the prober to recognize that no further retrieval is needed, allowing it to generate the correct answer.}
\vspace{-5mm}
\label{fig:intro}
\end{figure}

To optimize the RAG pipeline, recent research has introduced methods to adaptively adjust the number of retrievals.~\citep{su-etal-2024-dragin, mallen2022not, wang2023self}. 
A notable example is Adaptive-RAG~\citep{jeong-etal-2024-adaptive}, which classifies queries into three categories based on their complexity: no retrieval, single-step retrieval, or multi-step retrievals.
However, as shown in Figure~\ref{fig:intro}, external classifiers in Adaptive-RAG often fail to fully leverage the internal decision-making capabilities of the language model. This leads to unnecessary additional retrieval steps, resulting in knowledge conflicts between the model's internal knowledge and externally retrieved information~\cite{xie2024adaptive, xu2024knowledge}.
Other methods decide whether to retrieve based on token generation probabilities~\citep{jiang-etal-2023-active} and linguistic feedback~\citep{zhang-etal-2024-retrievalqa, ding2024retrieve} from the LLM.
However, these approaches still rely solely on the final output and cannot fully capture the model's internal reasoning processes.

In this paper, we propose Probing-RAG, which determines whether the model needs to retrieve documents to answer a given question by examining the internal representations of the language model.
As shown in the right example of Figure~\ref{fig:intro}, Probing-RAG focuses on the hidden states of the LLM's intermediate layers. 
These hidden states serve as inputs for the prober, which evaluates the need for further retrieval steps.
In essence, the prober assesses whether the language model can generate an answer to a query using the available information. This approach ensures reliable decisions regarding the necessity of additional retrieval.
We construct the prober by adding a fully connected layer to LLMs with a parameter size of just 5 MB, which is 2,000 times smaller than the external classification-based Adaptive-RAG~\cite{jeong-etal-2024-adaptive}.
To efficiently train the prober, we create a synthetic dataset derived from the widely used open-domain QA datasets.

We conduct experiments on five open-domain QA datasets to validate Probing-RAG. 
Experimental results demonstrate that Probing-RAG outperforms existing methods across both in-domain and out-of-domain datasets for the prober, reducing retrieval frequency by approximately 50\% on average.
Additionally, we provide a comprehensive analysis of prober training, including its position within the LLMs and the number of datasets required for training.

%
\section{Related Work}
\label{sec:preliminaries}

\paragraph{Adaptive Retrieval}
Adaptive retrieval methods~\citep{jiang-etal-2023-active, su-etal-2024-dragin, asai2023selfrag} have emerged as an innovative approach that allows language models to actively decide when and what external information to retrieve based on the task type or the specific information contained within the query.
These methods can be broadly categorized into three main approaches: external classifier based, LLM-based feedback, and confidence-based techniques.
\begin{figure*}[!ht]
\centerline{\includegraphics[width=\linewidth]{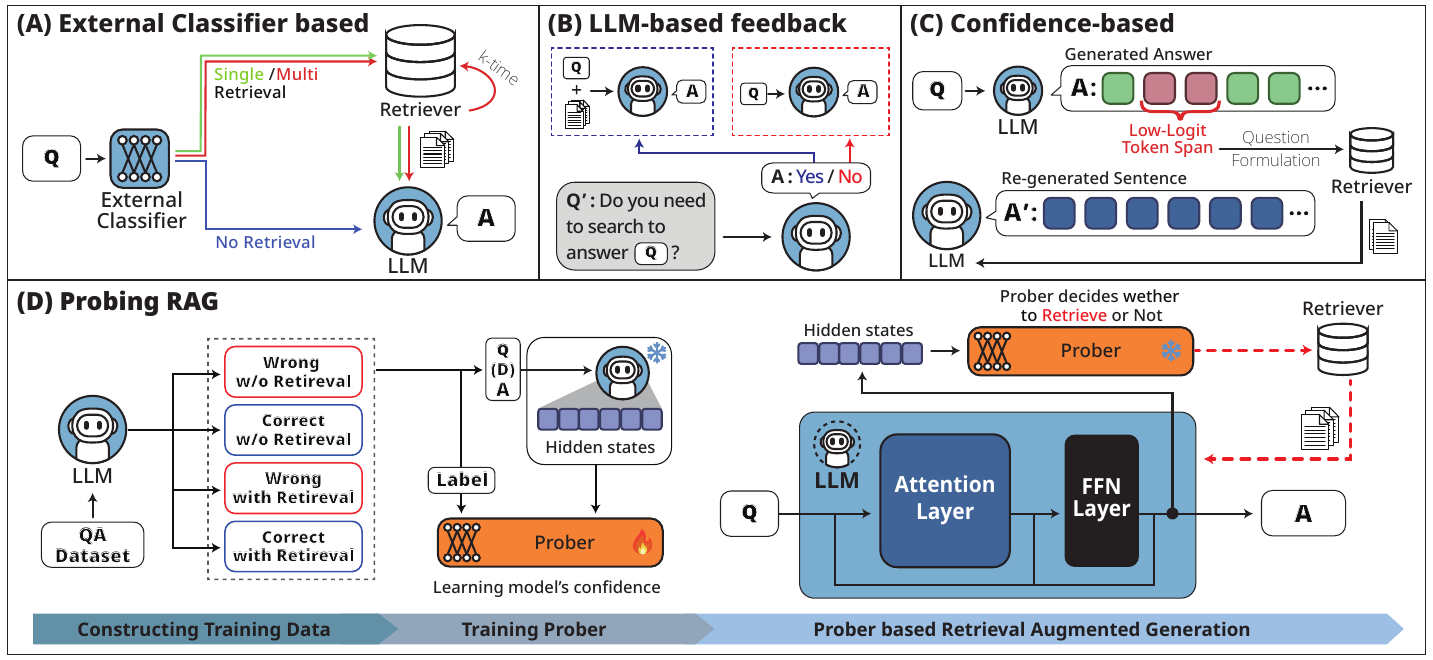}}

\caption{A conceptual comparison of various Adaptive-RAG approaches. (A) determines whether to perform retrieval based on query complexity measured by an external classifier. (B) decides retrieval based on the response from the LLM. (C) uses the confidence of the final token selection to determine retrieval. (D) Our proposed Probing-RAG decides retrieval using a prober model, which utilizes the internal hidden states of the LLM.}
\vspace{-4mm}
\label{fig:main}
\end{figure*}

External classifier based methods focus on training external models to optimize retrieval decisions based on query characteristics. 
For instance, Adaptive-RAG~\citep{jeong-etal-2024-adaptive} incorporates an additional classifier that categorizes queries into three types: no retrieval, single-step retrieval, or multi-step retrieval. This classifier selects the optimal number of retrievals based on the complexity of each query.
LLM-based feedback methods~\citep{zhang-etal-2024-retrievalqa, ding2024retrieve} rely on the evaluation of response consistency to guide retrieval decisions. These methods use prompting or multilingual formulations to assess the quality of the model's responses. 
If the LLM detects low consistency in its output, it triggers a retrieval to obtain relevant documents. 
Confidence-based approaches rely on the model's token uncertainty to guide retrieval decisions. For example, FLARE~\citep{jiang-etal-2023-active} initiates retrieval when any token in a generated sentence has a probability below a certain threshold, while DRAGIN~\citep{su-etal-2024-dragin} considers both token uncertainty and the attention weights between successive tokens to determine when to retrieve additional information.

Our work also aims to optimize RAG systems by dynamically adjusting the number of retrievals using a classifier similar to that of Adaptive-RAG. However, unlike Adaptive-RAG, which uses an external classifier, our study utilizes the internal representation of the model to effectively determine whether the model can solve the given query using the current context.

 \paragraph{Knowledge Conflict}
In RAG systems, knowledge conflicts~\citep{xu2024knowledge} can occur when external knowledge contradicts the parametric knowledge within LLMs, resulting in inconsistent outputs. Ideally, LLMs should identify these conflicts and provide distinct answers based on the conflicting contexts. However, LLMs often struggle to precisely detect and resolve such inconsistencies~\citep{wang2023resolving}. Instead, they often tend to overly rely on coherent and persuasive knowledge~\citep{xie2024adaptive}, which may not always reflect the most accurate or relevant information. This highlights the importance of using external knowledge only when the model is uncertain or lacks sufficient information.
 
\paragraph{Confidence Estimation via Internal State}
Estimating LLM’s factual confidence poses challenges when solely relying on outputs based on vocabulary distribution from the final layer~\citep{mahaut-etal-2024-factual}. To address this, studies have focused on the outputs of the feed-forward neural networks and attention heads in the intermediate layers of transformers. 
For summarization tasks, Lookback Lens~\citep{chuang2024lookback} measures confidence by comparing attention weights on context versus newly generated tokens.

Recent work has expanded beyond confidence estimation to address hallucination mitigation using intermediate representations. Dola~\citep{chuang2024dola} contrasts logits from later and earlier layers to obtain a more reliable next-token distribution. Lookback Lens mitigates contextual hallucinations using a classifier-guided decoding strategy trained on attention-weight ratios. Building on this line of work, we propose a method to dynamically trigger retrieval by employing a prober trained on the LLM's hidden states.

\section{Method}
\label{sec:method}

We introduce Probing-RAG, an efficient RAG pipeline that incorporates a prober to determine whether the language model needs to retrieve additional documents.
Similar to the conventional retrieval-augmented generation pipeline, our approach comprises a generating language model and a retriever.
Different from the general pipeline, the generator of Probing-RAG leverages the output from the prober and adaptively calls the retriever based on the model's internal hidden state.

\subsection{Prober}
Given the LLM's hidden state during answer generation, the prober assesses whether an additional retrieval step is necessary.
We design the prober as a feed-forward network with a single hidden layer and an output layer for binary classification.
Based on the findings of \citet{chuang2024dola}, which indicate that lower layers in language models capture low-level information while higher layers capture more abstract, high-level information, we position the prober after the one-third point of the model to maximize the utility of these representations. 
In our experiments, we employ Gemma-2B~\citep{team2024gemma}, an 18-layer model, as the generator in our RAG pipeline. Consequently, we position the prober on even-numbered layers starting from the 6th layer.

The prober utilizes the hidden states corresponding to the model-generated rationale ($r$) and answer ($\hat{a}$). Let $H_{l_k}$ represent the hidden state at the $k$th layer, and let the combined length of the $r$ and $\hat{a}$ tokens be $u$. We define the hidden states corresponding to these tokens as:

\begin{equation}
    \resizebox{0.30\textwidth}{!}{
    $T_{l_k} = H_{l_k}[-u:, d_{\text{model}}] \in \mathbb{R}^{u \times d_{\text{model}}}$
    }
\end{equation}

To reduce the dimensionality of $T'_{l_k}$ and obtain a summary representation, we first compute the mean of the hidden states across the token dimension, producing a single vector. Next, we normalize this vector to maintain numerical stability and consistency, resulting in the final input to the prober:

\begin{equation}
    \resizebox{0.37\textwidth}{!}{
    $T'_{l_k} = \text{Norm}(\text{Mean}(T'_{l_k}, \text{dim}=0)) \in \mathbb{R}^{d_{\text{model}}}$
    }
\end{equation}

This normalized representation $T'_{l_k}$ is then fed into the prober:

\begin{equation}
    \resizebox{0.22\textwidth}{!}{
    $\text{logit}_{l_k} = \text{Prober}_{l_k}(T'_{l_k})$
    }
\end{equation}

Using the value of $\text{logit}_{l_k}$, prober decides whether or not to perform an additional retrieval.

\begin{figure}[!ht]

\centerline{\includegraphics[width=\linewidth]{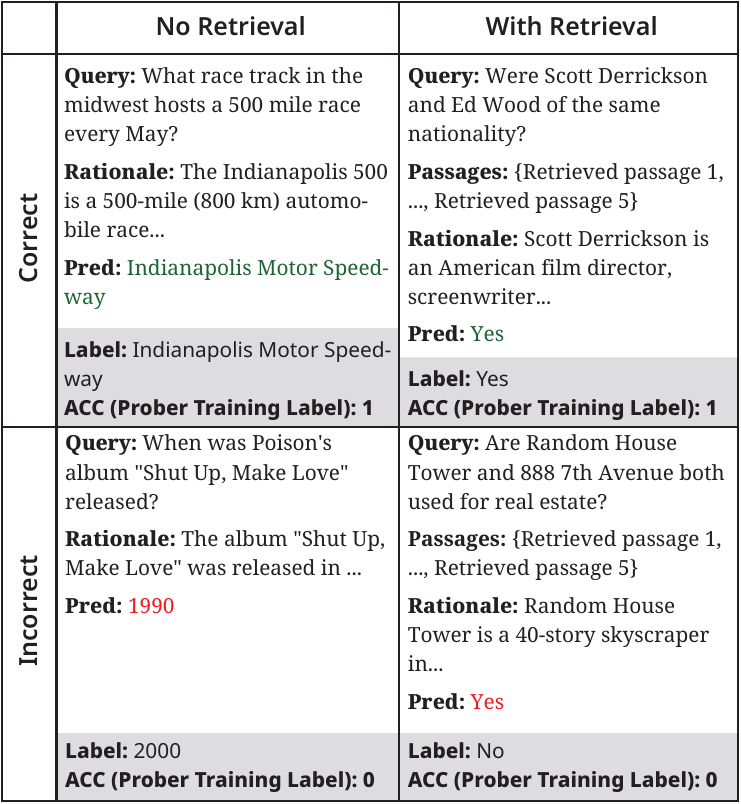}}
\caption{Examples of prober training dataset.}
\vspace{-5mm}
\label{fig:datasets}
\end{figure}
\subsection{Training Prober}
\label{sub:method_data}

To train the prober, we require pairs of $(T'_{l_k}, y)^N_{1}$, where $T'_{l_k}$ is the input derived from the hidden states, and $y$ is the output denoting whether additional retrieval is needed. 
To generate these pairs, we first use Chain-of-Thought (CoT)~\citep{wei2022chain} prompting to let the LLM generate $r$ and $\hat{a}$ for the given question. 
The hidden state, $T'_{l_k}$, is recorded during this answer generation, and the label $y$ is assigned based on the accuracy of $\hat{a}$ compared to the ground truth. 
If the answer is correct, $y = 1$ (indicating no retrieval is needed); if the answer is incorrect, $y = 0$ (indicating retrieval is necessary).

For each question ($q$), two versions of $r$ and $\hat{a}$ are generated: one without retrieval and one with retrieval. This dual approach helps the prober learn to distinguish when retrieval improves performance and when it does not. By comparing hidden state representations across both retrieval and non-retrieval cases, the prober can better detect situations where retrieval is likely to improve accuracy. As shown in Figure~\ref{fig:datasets}, the dataset includes four types of examples: (with retrieval, $y = 0$), (with retrieval, $y = 1$), (without retrieval, $y = 0$), and (without retrieval, $y = 1$). 

We use three open-domain QA datasets to build the training dataset: HotpotQA~\citep{yang-etal-2018-hotpotqa},  NaturalQA~\citep{kwiatkowski-etal-2019-natural}, and TriviaQA~\citep{joshi-etal-2017-triviaqa}. 
To create a balanced dataset, we ensure an equal distribution of correctly and incorrectly answered questions.
The final dataset consists of 26,060 training and 500 validation samples.

\begin{algorithm}[!t]
    \footnotesize	
    \caption{Probing based RAG}
    \textbf{Input}: \( Q \) (question), \( D \) (corpus), \( d_j \) (retrieved jth document) \(\mathcal{M}\) (LLM), \(R\) (retriever)\\ 
    \textbf{Initialize}: set $\text{count} = 0$, $Logits=\phi$
    \begin{algorithmic}[1]
    \State Define $H_{l_k} =$ the hidden state at the pre-defined position of the k-th layer.
    \State Define $r , \hat{a}=$ the rationale and predicted answer tokens generated by $\mathcal{M}$.
    \State Define $L, l_{k} =$ the set of layers and each individual layer of $\mathcal{M}$.
    \State Define $\theta =$ the threshold.  
    \State $(r, \hat{a})\leftarrow\mathcal{M}.generate(q)$
        \While {count $<$ max iteration}
            \If {count is not 0}
            \State $(r, \hat{a}) \leftarrow \mathcal{M}.generate(q, d_1, \ldots, d_j)$
            \EndIf
            \State $Logits \leftarrow \phi$
            \For {$L \in \{l_6, l_8, \ldots, l_k\}$}
            \State $u = len(r) + len(\hat{a})$
            \State $T_{l_k} \leftarrow H_{l_k}([-u:,d_{model}])$ 
            \State $T'_{l_k} \leftarrow normalize (mean(T_{l_k}, dim = 0))$
            \State $logit_{l_k} \leftarrow Prober_{l_k} (T'_{l_k})$
            \State Add $logit_{l_k}$ to $Logits$
            \EndFor
            \State call retrieval, pass retrieval $\leftarrow Logits[0], Logits[1]$
            \If{call retrieval $+ \theta$ > pass retrieval}
            \If{count is 0}
            \State $d_1, \ldots, d_j \leftarrow R(q, r, \hat{a})$
            \Else
            \State $d_1, \ldots, d_j \leftarrow R(q, d_1, \ldots, d_j,r, \hat{a})$
            \EndIf
            \State $\text{count} \leftarrow \text{count} + 1$
            \Else
            \State break
            \EndIf
        \EndWhile

    \end{algorithmic}
    \textbf{Output}: $\hat{a}$.
    \label{alg:probing}
\end{algorithm}

\label{sub:method_train}
We train the probers using the generated dataset with a cross-entropy loss as follows:
\begin{equation}
    \resizebox{0.42\textwidth}{!}{
    $L = - \frac{1}{N} \sum_{i=1}^{N} \left[y_i \log(p_i) + (1 - y_i) \log(1 - p_i)\right]$
    }
\end{equation}

We present the hyperparameter details for training the prober in Appendix~\ref{appen:hyper}.

\subsection{Probing based Retrieval-Augmented Generation}
\label{sub:method_inference}
In the first iteration, using CoT prompting, the LLM generates $r$ and $\hat{a}$ based solely on the input question $q$:
\begin{equation}
    \resizebox{0.43\textwidth}{!}{
    $(r, \hat{a}) = 
    \begin{cases} 
    LLM.generate(q) & \text{if iteration = 0}\\
    LLM.generate(q,\{d_k\}_1^j) & \text{otherwise}
    \end{cases}$}
\end{equation}
After generating the initial $r$ and $\hat{a}$, the prober assesses whether retrieval is necessary. To do this, we extract hidden state representations $T'_{l_k}$ of the $r$ and $\hat{a}$ tokens from intermediate layers and feed them into the probers assigned to each layer to generate logit values. 
The decision to perform additional retrieval is based on the sum of the probers' logit values.
If the difference between the logit for retrieval necessity and the logit indicating no need for retrieval is higher than the threshold $\theta$, additional documents ${d_k}_1^j$ are retrieved based on the following conditions:
\begin{equation}
    \resizebox{0.43\textwidth}{!}{
    $\{d_k\}_1^j=
    \begin{cases} 
    R(q, r, \hat{a}) & \text{if iteration = 0}\\
    R(q, \{d_k\}_1^j, r, \hat{a}) & \text{otherwise}
    \end{cases} $
    }
\end{equation}
In subsequent iterations, the LLM uses these retrieved documents along with $q$ to generate updated versions of $r$ and $\hat{a}$. 
This iterative process continues until either no further retrieval is needed or the maximum number of iterations is reached. 
Detailed information on procedure of the Probing-RAG is provided in Algorithm~\ref{alg:probing}.
\begin{table*}[!ht]
\small
\resizebox{\linewidth}{!}{\begin{tabular}{l||cccccc||cccc||cc}

\toprule
\multirow{3}{*}{Methods} & \multicolumn{6}{c||}{In-Domain}& \multicolumn{4}{c||}{Out-of-Domain} & \multicolumn{2}{c}{\multirow{2}{*}{Average}}\\
 & \multicolumn{2}{c}{HotpotQA} & \multicolumn{2}{c}{NQ}& \multicolumn{2}{c||}{TriviaQA}&\multicolumn{2}{c}{MuSiQue}&\multicolumn{2}{c||}{2Wiki} & \\ \cmidrule(lr){2-13}
& EM & ACC & EM & ACC & EM & ACC & EM & ACC & EM & ACC & EM & ACC \\ \midrule
Gemma-2b &  & &  &  & &&&&&&&\\
\quad No Retrieval & 16.8 & 28.0& 15.0 & 24.6 & 37.4& 45.4 & 3.2& 4.8 &22.6&\underline{43.0}&19.0&\underline{29.2}\\
\quad Single-step Approach & 14.6 & \underline{28.2} & 11.4 & 26.0 & 19.6 & 38.8 & 1.8 & \underline{5.8} &22.8&38.4&14.0&27.4\\

\midrule
\quad LLM-based& 18.6 & 25.8 & 17.6 & 20.4 & 36.8 & 41.8 & 3.8 & 4.6 & 24.2 & 25.8 & 20.2 & 23.7 \\
\quad FLARE& 13.2 & 21.0 & 9.0& 21.8 & 13.8 & 31.0 & 1.2 & 5.0 &21.6&27.8&11.8&21.3\\
\quad DRAGIN& \underline{19.8} & 22.6 & \underline{18.8}& 22.2 & \textbf{42.8} & \underline{47.0} & \underline{4.2} & 4.8 &\textbf{26.4}&28.8&\underline{22.4}&25.1\\
\quad Adaptive-RAG & 13.2 & 23.6 & 11.4 & \underline{26.2} & 22.8 & 40.8 & 1.2 & 3.0&21.6&40.6&14.0&26.8\\

\quad Probing-RAG(Ours)& \textbf{21.8} & \textbf{39.4} & \textbf{21.6} & \textbf{35.0} & \underline{41.8} & \textbf{52.2} & \textbf{4.8} & \textbf{8.8} & \underline{24.2} & \textbf{43.6} & \textbf{22.8} & \textbf{35.8} \\
\midrule
\midrule
Mistral-7b &  & &  &  & &&&&&&&\\
\quad No Retrieval & 17.0&20.6&13.2&19.8&38.0&45.2&3.4&6.2&16.4&30.0&17.6&24.4\\
\quad Single-step Approach & 18.6&\underline{34.2}&16.8&35.0&34.6&\underline{51.0}&5.4&9.0&21.6&\underline{32.6}&19.4&\underline{32.4}\\

\midrule
\quad LLM-based& 20.4&32.0&15.8&35.6&41.2&49.8&\underline{5.6}&\underline{9.4}&16.8&32.4&20.0&31.8\\
\quad FLARE& 20.4&32.0&15.4&34.4&35.0&45.6&4.4&6.6&18.6&31.0&18.8&29.9\\
\quad DRAGIN& \underline{21.2}&28.0&16.8&37.2&39.8&42.2&5.2&7.2&\textbf{23.2}&25.8&\underline{21.2}&28.1\\
\quad Adaptive-RAG & 19.0&26.0&\underline{17.2}&\underline{37.4}&\underline{40.8}&50.2&4.0&5.8&22.6&31.6&20.7&30.2\\

\quad Probing-RAG(Ours)& \textbf{22.4}&\textbf{38.6}&\textbf{20.8}&\textbf{39.4}&\textbf{43.2}&\textbf{52.2}&\textbf{5.8}&\textbf{9.8}&\underline{23.0}&\textbf{33.4}&\textbf{23.0}&\textbf{34.7} \\
\bottomrule
\end{tabular}}
\caption{Experimental results on five different open-domain QA datasets. We indicate the highest performance in bold and underline the second highest.}
\vspace{-3mm}
\label{tab:main}
\end{table*}

\section{Experiment}
\subsection{Experimental Setup}

\paragraph{Datasets}
We use five open-domain QA datasets for our experiments. We split the datasets used for training the prober as the in-domain dataset, while those not used for training are considered the out-of-domain dataset.
The in-domain datasets include NaturalQA (NQ)~\cite{kwiatkowski-etal-2019-natural}, TriviaQA~\cite{joshi-etal-2017-triviaqa}, and HotpotQA~\cite{yang-etal-2018-hotpotqa}, and the out-of-domain datasets consist of MuSiQue~\cite{trivedi-etal-2022-musique} and 2WikimultihopQA (2Wiki)~\cite{ho-etal-2020-constructing}. To evaluate the performance of each Adaptive-RAG method, we sample 500 examples from the test set of each dataset.

\paragraph{Baselines} 
We use the following baselines to evaluate the performance: \textbf{No Retrieval}, which proceeds directly based on the query without retrieval. 
\textbf{Single-step Approach}, which conducts a single retrieval call before QA. 
The \textbf{LLM-based} approach determines whether retrieval is necessary based on the language model’s response to the question.
\textbf{FLARE}~\citep{jiang-etal-2023-active} is a multi-round retrieval-augmented method that triggers retrieval each time it encounters an uncertain token. When a sentence contains uncertain tokens, a query is generated to retrieve relevant passages. The sentence is then replaced with a newly generated one based on the query and the retrieved passage. 
\textbf{DRAGIN}~\citep{su-etal-2024-dragin} determines when to retrieve based on token generation probabilities and performs query reformulation using attention weights. 
\textbf{Adaptive RAG}~\citep{jeong-etal-2024-adaptive} performs by fine-tuning a query complexity classifier.
All methods use 4-shot examples, and we evaluate performance using Exact Match (EM) and Accuracy (ACC).

\paragraph{Implementation Details}
We use BM25~\citep{robertson1976relevance, robertson2009probabilistic}, a term-based sparse retrieval model, for all frameworks to ensure a fair comparison. 
We use the same document corpus and datasets as~\citet{jeong-etal-2024-adaptive}.
For Adaptive-RAG, we train the classification model using the t5-large~\citep{chung2024scaling} model, following the approach of \citet{jeong-etal-2024-adaptive}.
We use the Gemma-2b~\citep{team2024gemma} and Mistral-7b~\citep{jiang2023mistral} model as the QA model for all frameworks. 
We use the residual post (epoch=2) as the prober’s hidden state representation position for Probing-RAG training and evaluation.

\subsection{Main Results}
\begin{table}[t]
\resizebox{\linewidth}{!}{\begin{tabular}{l||c||ccc}
\toprule
   & \multirow{2}{*}{Total Retrieval Call} & No    & Single-step & Multi-step \\ \cline{3-5} 
   &             & \multicolumn{3}{c}{Retrieval Step Ratio} \\ \midrule
LLM-based         &   2345  & 6.2\% & 93.8\%   & 0.00\%  \\
FLARE & 5317            & 12.41\% & 29.35\%   & 58.24\%  \\
DRAGIN & 13570            & 0.00\% & 1.20\%   & 98.80\%  \\
Adaptive-RAG & 3068            & 7.79\% & 61.96\%   & 30.25\%  \\
Probing-RAG  & 1988            & 57.46\% & 20.19\%   & 22.35\%   \\ \bottomrule
\end{tabular}}
\caption{Distribution of retrieval steps using the Gemma-2b model, showing the proportion of instances with no, single-step, and multi-step retrieval.}
\vspace{-5mm}
\label{tab:cls}
\end{table} 
We present the performance of each method on open-domain QA tasks in Table~\ref{tab:main}. We first observe that the single retrieval results outperform the no retrieval results in accuracy across all cases, except for TriviaQA and 2WikimultihopQA. We attribute the performance degradation when using the single-step approach on TriviaQA and 2WikimultihopQA to irrelevant external knowledge causing the decrease. 
We demonstrate that Probing-RAG demonstrates the best performance, with improvements of approximately 6.59\% points and 8.35\% points in ACC compared to the no-retrieval and single-step approaches, respectively.
We also observe distinct characteristics in the results for each Adaptive-RAG method. 
We find that the FLARE method demonstrates lower performance compared to no retrieval across all datasets except for MuSiQue, and performs worse than the single-step approach across all datasets. 
For the FLARE, it continuously makes LLM calls to generate new queries and sentences that replace low-probability tokens, demonstrating its heavy reliance on model-generated text and effectiveness primarily in large-parameter models.
The DRAGIN shows lower accuracy compared to the single-step approach except for TriviaQA. However, as shown in Table~\ref{tab:cls}, DRAGIN generally performs over five retrievals, which enhances exact match scores. 
We discuss DRAGIN’s performance in more detail in the case study in Table~\ref{tab:case} and in Appendix~\ref{appen:em}.
Next, the Adaptive-RAG method outperforms the single-step approach on NQ, TriviaQA, and 2WikimultihopQA but shows lower results on the other datasets. Adaptive-RAG relies on an externally trained model to access query complexity, which fails to account for the internal knowledge of the QA model. 
We demonstrate that our proposed Probing-RAG outperforms all of these previous adaptive retrieval methods by avoiding redundant retrieval. By leveraging hidden state representations of LLMs, our approach leads to significant performance improvements.

\subsection{Analysis}
\paragraph{Number of Retrieval Steps}
In Table~\ref{tab:cls}, we present the total number of retrievals for each method along with the corresponding retrieval step ratio.
We observe that LLM-based, FLARE, Adaptive-RAG, and DRAGIN perform 1.17, 2.67, 1.54, and 6.83 times more retrieval calls, respectively, compared to Probing-RAG. 
Probing-RAG performs retrieval calls when necessary, which is also reflected in the retrieval step ratio of Table~\ref{tab:cls}. 
FLARE and Adaptive-RAG skip retrieval for 12.41\% and 7.79\% of cases and execute at least one retrieval for the rest.
Moreover, DRAGIN performs multi-step retrieval in 98.8\% of cases, resulting in computational overhead. However, as shown in Table~\ref{tab:main}, ACC performance is lower than Probing-RAG. Our method does not perform retrieval in 57.46\% of cases. 
This demonstrates that the model successfully answers a higher proportion of queries without relying on retrieval, indicating that unnecessarily increasing the number of retrieval calls leads to computational overhead without effectively enhancing performance.

\begin{table}[t]
\centering
\resizebox{0.9\linewidth}{!}{\begin{tabular}{l||ccc}
\toprule
 & In-domain & Out-of-domain& Total\\ \midrule
\multicolumn{1}{l||}{FLARE} & 24.55 & 16.40 & 21.28 \\
\multicolumn{1}{l||}{DRAGIN} & 30.61 & 16.77 & 25.07 \\
\multicolumn{1}{l||}{Adaptive-RAG} & 30.15 & 21.76 & 26.79 \\ \midrule
\multicolumn{1}{l||}{Probing-RAG} & 42.18 & 26.15 & 35.77\\
\multicolumn{1}{l||}{\quad 1/2 of Train Set} & 41.32 & 25.35 & 34.93\\
\multicolumn{1}{l||}{\quad 1/4 of Train Set} & 40.65 & 26.35  & 34.93 \\
\multicolumn{1}{l||}{\quad 1k Samples}  & 40.32 & 25.75 & 34.49\\
\bottomrule
\end{tabular}}

\caption{Performance comparison of variations in training data size for Probing-RAG and different retrieval methods using the Gemma-2b model. Values shown are accuracy (\%) for in-domain, out-of-domain, and total averages.}
\vspace{-4mm}
\label{tab:size}
\end{table}

\begin{figure*}[!ht]
\centerline{\includegraphics[width=1.0\linewidth]{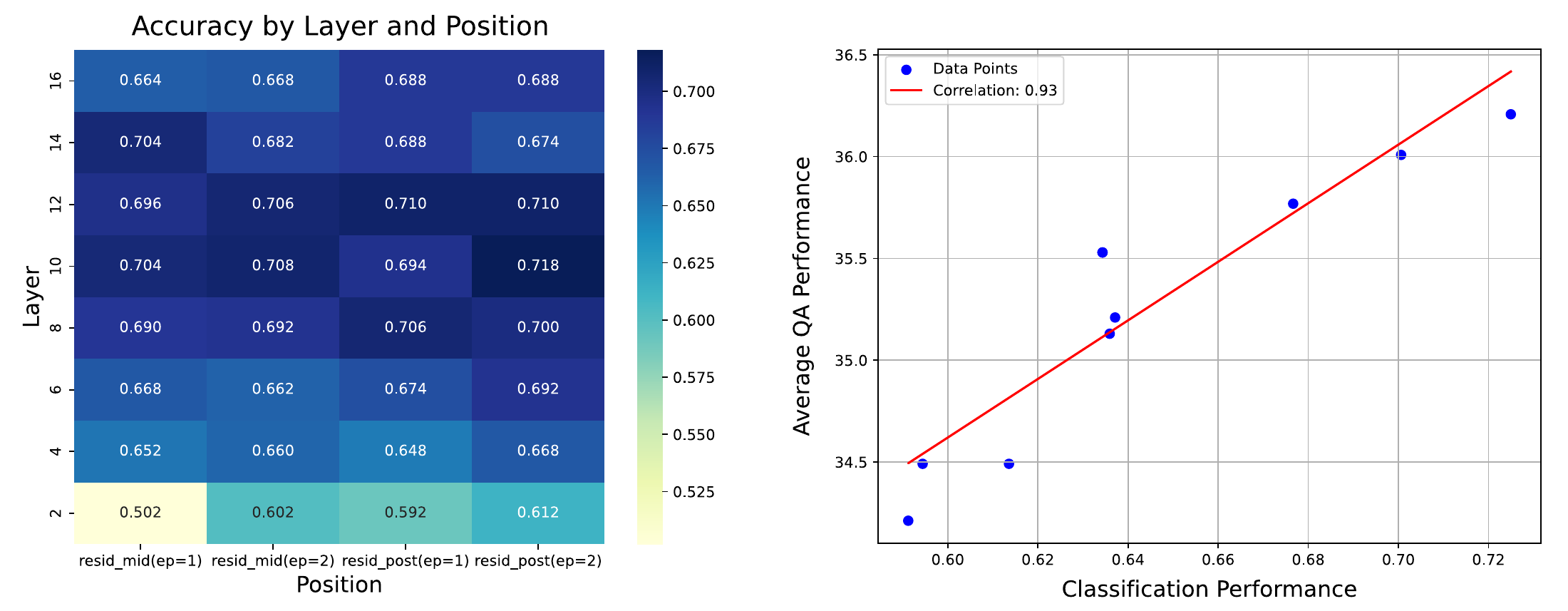}}
\caption{(Left) Probing accuracy measured for each model among the layer, (Right) Correlation between prober's classification performance and QA performance, using the Gemma-2b model.}

\label{fig:heat}
\end{figure*}
\paragraph{Size of Prober Training dataset}
To analyze the impact of training data size on performance in our framework, we conduct experiments by training the prober with varying amounts of data.
We measure the performance on open-domain QA by training the prober on reduced datasets consisting of a sample of 1,000 data, as well as 1/4 and 1/2 proportions of the total 26,060 data points we present the result in Table~\ref{tab:size}.
As expected, the overall results show that in most cases, performance improves as the amount of training data increases. However, we also observe that the model trained with only 1k data points is comparable to the prober trained on one-third of the entire dataset in most of the datasets. 
Furthermore, we find that the prober trained on just 1k data points outperforms all of the previous methods, indicating that  that it is possible to effectively train the prober using a small dataset.
\paragraph{Answer Consistency among Retrieval Usage}
We define model consistency as the ability of a model to correctly answer queries with retrieval methods that it previously answered correctly without retrieval. 
Specifically, low model consistency indicates situations where the model answers incorrectly due to the intervention of external knowledge, despite being capable of answering the query using only its internal knowledge. This phenomenon highlights the potential drawbacks of incorporating external information for queries that the model could originally handle independently. 
In Table~\ref{tab:consistency}, we evaluate how consistently single-step RAG, FLARE, DRAGIN, Adaptive-RAG, and Probing-RAG answer queries solvable without retrieval. 
We observe that Probing-RAG achieves the highest consistency across all four datasets, each consisting of 500 examples. In contrast, as shown in Table 2, single-step RAG, FLARE, DRAGIN, and Adaptive-RAG perform retrieval for most queries. This leads to significant drops in consistency because incorrect retrieval timing and unnecessary external information cause the models to generate inconsistent answers.
\begin{table}[t]
\resizebox{\linewidth}{!}{
\begin{tabular}{l||ccccc}
\toprule
Consistency& HotpotQA   & NQ & TriviaQA & MuSiQue &2Wiki \\ \midrule

\multicolumn{1}{l||}{No Retrieval}   & 100\%  & 100\%  & 100\%   & 100\% &100\%\\
\multicolumn{1}{l||}{Single-step-RAG}   & 73.8\%  & 76.0\%           & 82.0\%   & 70.4\% &85.0\% \\
\multicolumn{1}{l||}{FLARE}   & 74.4\%  & 76.4\%           & 83.8\%   & 68.8\% &72.8\%  \\
\multicolumn{1}{l||}{DRAGIN}   & 75.2\%    & 75.0\%           & 81.8\%   & 77.8\% &69.8\% \\
\multicolumn{1}{l||}{Adaptive-RAG} & 83.0\%   & 76.6\%           & 86.0\%   & 74.6\% &93.2\%\\
\multicolumn{1}{l||}{Probing-RAG}  & 90.6\%   & 92.6\%           & 91.0\%   & 96.4\% &96.6\% \\ \bottomrule
\end{tabular}
}
\caption{Comparison of accuracy between no retrieval and adaptive retrieval augmentation to evaluate model consistency using the Gemma-2b model.}

\label{tab:consistency}
\end{table}


\subsection{Prober Accuracy}
\paragraph{Accuracy among Layers}
We validate the performance of prober in classifying the necessity of retrieval calls using the test set. In Figure~\ref{fig:heat}, the left side shows accuracy according to position and layer. 
Most probers achieve a performance near 0.7, with the highest performance observed at the residual post position. Therefore, we use the prober at this position for performance evaluation.
\begin{table*}[]
\resizebox{\linewidth}{!}{\begin{tabular}{l|l|l}
\toprule
\multicolumn{1}{c|}{\textbf{Question \& Label}}& \multicolumn{1}{c|}{\textbf{Probing-RAG}} & \multicolumn{1}{c}{\textbf{DRAGIN}} \\
\midrule
\begin{tabular}[c]{@{}l@{}}\textbf{Question 1:}\\ Ezzard Charles was a world champion \\in which sport? \\ \textbf{Label:}\\ 'Fistfighting', 'Corner men', 'Corner persons',\\ 'Prize-fighter’, Boxing’ ...  \end{tabular} 
& \begin{tabular}[c]{@{}l@{}}\textbf{Rationale:} Ezzard Charles was a world champion \color{blue}{boxer}, \\ winning the world heavyweight title in 1949. \\He was born in ... \\ \textbf{Answer:} Boxing \\ \\ \textbf{ACC:} 1, \textbf{EM:} 1\\ \textbf{Total Number of Retrieval call:} 0\end{tabular} 
& \begin{tabular}[c]{@{}l@{}}\textbf{Rationale:} Ezzard Charles was a world champion \\in \color{blue}{boxing}. \\ \textbf{Answer:} Boxing\\ \\ \textbf{ACC:} 1, \textbf{EM:} 1\\ \textbf{Total Number of Retrieval call:} 8\end{tabular} \\ 
\midrule

\begin{tabular}[c]{@{}l@{}}\textbf{Question 2:}\\Alfred Balk served as the secretary of \\the Committee on the Employment of Minority \\Groups in the News Media under which \\United States Vice \\ \textbf{Label:}\\ 'Nelson Rockefeller' \end{tabular} 
& \begin{tabular}[c]{@{}l@{}}\textbf{Rationale:} The Committee on the Employment of \\Minority Groups in the News Media was established \\by the United States Vice President \color{blue}{Nelson Rockefeller}\\ in 1972... \\ \textbf{Answer:} Nelson Rockefeller \\ \\ \textbf{ACC:} 1, \textbf{EM:} 1\\ \textbf{Total Number of Retrieval call:} 3\end{tabular} 
& \begin{tabular}[c]{@{}l@{}}\textbf{Rationale:} The National News Council was established \\ by the Twentieth Century Fund in 1965. The committee \\ was established to study... \\ \textbf{Answer:} \color{red}{Lyndon B. Johnson}\\ \\ \textbf{ACC:} 0, \textbf{EM:} 0\\ \textbf{Total Number of Retrieval call:} 8\end{tabular} \\ 
\midrule

\begin{tabular}[c]{@{}l@{}}\textbf{Question 3:} \\ Leslie Nielsen trained in which of \\the armed services in WWII? \\ \textbf{Label:} \\ 'Royal Canadian' \end{tabular} 
& \begin{tabular}[c]{@{}l@{}}\textbf{Rationale:} Leslie Nielsen was a Canadian actor, best \\known for his role as Dr. Strangelove in the 1964... \\ \textbf{Answer:} \color{orange}{Royal Canadian Air Force} \\ \\ \textbf{ACC:} 1, \textbf{EM:} 0\\ \textbf{Total Number of Retrieval call:} 0\end{tabular} 
& \begin{tabular}[c]{@{}l@{}}\textbf{Rationale:} Leslie Nielsen trained in the \color{blue}{Royal Canadian}.\\ \textbf{Answer:} Royal Canadian\\ \\ \textbf{ACC:}1, \textbf{EM:} 1\\ \textbf{Total Number of Retrieval call:} 6\end{tabular} \\ \bottomrule

\end{tabular}}

\caption{A case study using the Probing-RAG and DRAGIN methods. Accurate information referenced in the response is marked in \textcolor{blue}{blue}, factual errors in \textcolor{red}{red}, and responses that are correct but not an exact match in \textcolor{orange}{orange}.}

\label{tab:case}
\end{table*}

\paragraph{Correlation between Prober Accuracy and QA Performance}
We analyze the prober's effectiveness by demonstrating the correlation between its classification performance and the average performance across five open-domain QA datasets, and present the results on the right of Figure~\ref{fig:heat}.
We utilize the prober classification and average performance obtained from varying thresholds in Table~\ref{tab:threshold}, along with those based on the number of probing layers in Table~\ref{tab:ablation}. 
As Figure~\ref{fig:heat} illustrates, improvements in the prober's performance correspond to increases in the average, as indicated by a strong correlation of 0.93. 
Through this experiment, we demonstrate the effectiveness of the Probing-RAG approach, which utilizes trained probers that take the model's internal hidden states as input.

\begin{figure}[t]
\centerline{\includegraphics[width=0.75\linewidth]{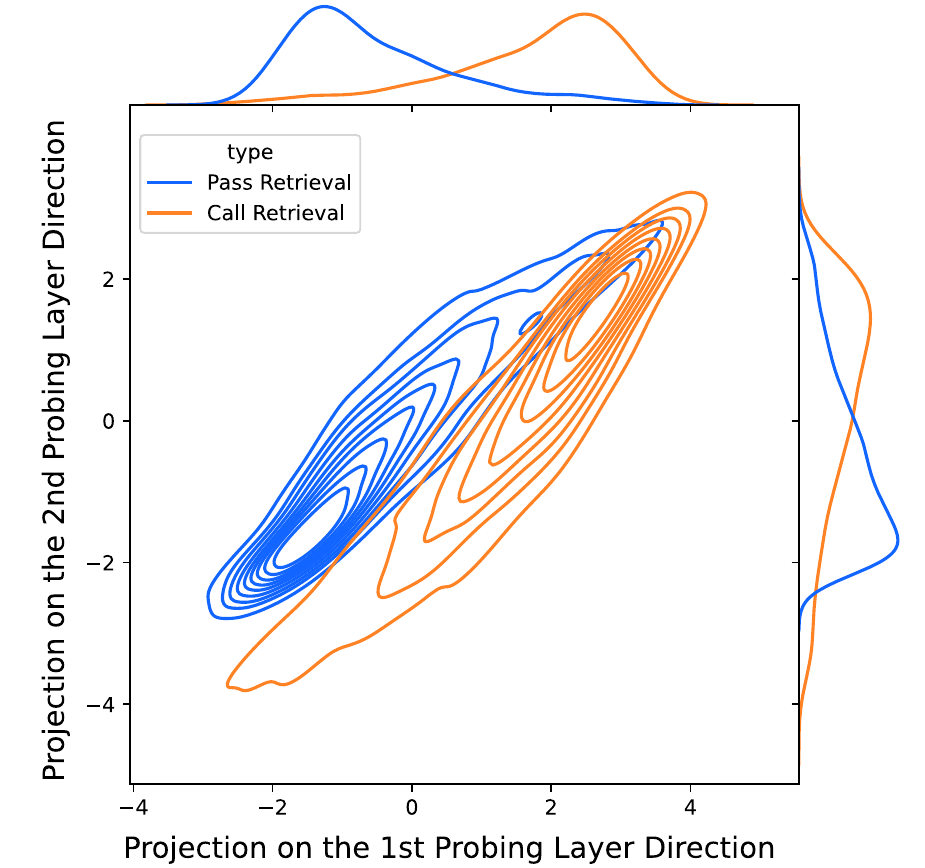}}
\caption{Kernel density estimate plot of logits using the Gemma-2b model, where orange indicates retrieval calls and blue indicates no retrieval needed. Marginal distributions are shown on the top and right. The results are projected onto the 10th and 12th residual post positions.}
\vspace{-5mm}

\label{fig:kde_}
\end{figure}

\paragraph{Validity of Prober Training}
To demonstrate the validity of the prober trained on our constructed dataset and hidden states, we conduct kernel density estimation experiments. For the experiment, we extract the logit values of the prober based on the 500 test datasets created in Section~\ref{sub:method_data}. Additionally, we use the logits of the two best-performing probers at the residual post to create distributions for scenarios when retrieval is used and when it is not. In Figure~\ref{fig:kde_}, the overlap between the distributions of the generated logits is minimal; the logit values where retrieval is used are distributed toward the upper right, while those that do not require additional retrieval are distributed toward the lower left. This demonstrates that the prober, trained on the structure of our dataset as well as the hidden states of the rationale and answer tokens, effectively classifies the necessity of retrieval calls.

\subsection{Case Study}
We conduct a case study comparing the Probing-RAG and DRAGIN methods across three QA pairs from the TriviaQA and NQ datasets, with the results presented in Table~\ref{tab:case}. 
We measure the accuracy, exact match, total retrieval calls, and the rationale and answers generated by the model. 
In Question 1, Probing-RAG only uses the model's internal knowledge to generate an answer without any retrieval. 
On the other hand, DRAGIN makes eight retrievals, even for questions that can be answered using the model's internal knowledge, composes a single-sentence rationale, and then generates the final answer. 
For the Question 2, the prober determines that retrieval is necessary, conducts three retrieval steps, extracts the rationale, and provides an answer. 
However, the DRAGIN method performs eight retrievals but generates a completely different answer. We demonstrate that repeated retrieval calls do not necessarily enhance performance and negatively affect it due to knowledge conflicts. 
In the final Question 3, our method provides the correct answer without retrieval but receives an exact match score of zero. 
In contrast, DRAGIN generates a concise single sentence and produces an accurate answer. Our analysis reveals that DRAGIN performs retrieval more than five times on average, generates a brief one-sentence rationale when forming answers, and achieves high exact match performance. This indicates that DRAGIN's heavy reliance on external knowledge during answer generation contributes to improvements in exact match performance. 
In Appendix~\ref{appen:em}, we provide a detailed comparison between ProbingRAG and DRAGIN on key factors such as the length of the generated answers, the number of retrieval calls, and the reasons for its high exact match scores.
\section{Conclusion}
In this study, we introduce Probing-RAG, an efficient adaptive retrieval pipeline that uses a pre-trained prober to determine whether additional retrieval is necessary by utilizing the language model's hidden states. 
We introduce both the training dataset and the training method for the prober and show that Probing-RAG outperforms previous methods for various open-domain QA datasets. 
We also show that Probing-RAG reduces overhead through appropriate retrieval calls by effectively leveraging both external and internal knowledge via the prober.
\section*{Limitations}
We propose a Probing-RAG method that reduces overhead by deciding whether to use retrieval based on the model’s hidden states and achieves high performance. However, the Probing-RAG method is limited to open-source LLMs, as it is incompatible with certain APIs where hidden state access is restricted. 
Additionally, our method involves elaborate data generation and some computational costs for training the prober. However, we also demonstrate that effective prober training is feasible with a small dataset.
Furthermore, due to resource constraints, we were unable to test the prober on a broader range of models, including those with hyper-scale models such as 70B parameters, or validate its effectiveness on domain-specific datasets, which may limit its generalizability.
\section*{Ethics Statement}
This study conducts QA tasks in the field of retrieval-augmented generation using language models. It also involves data generation and training. Therefore, it is important to recognize that language models may produce inappropriate responses. Additionally, as it may retrieve inappropriate content from the searched documents, developing management methods for this is essential. We believe this is a crucial area for future work.

\section*{Acknowledgement}
This research was supported by Institute for Information \& Communications Technology Planning \& Evaluation (IITP) through the Korea government (MSIT) under Grant No. 2021-0-01341 (Artificial Intelligence Graduate School Program (Chung-Ang University)).

\bibliography{acl}

\appendix

\appendix
\newpage
\clearpage
\section{Appendix}
\subsection{Hyperparameters}
\label{appen:hyper}
\begin{table}[hbt]
  \centering
  {
    \begin{tabular}{l|c}
      \toprule
      \textbf{Name}                    & \textbf{Value}  \\
      \midrule
      Learning rate & 1e-3    \\
      Batch & 12 \\
      Epoch      & 2   \\
      Dropout & 0.1\\
      Activation function & Silu\\
      Noramlize & LayerNorm \\
      Optimizer     & AdamW \\
      scheduler     & ExponentialLR \\
      gamma & 0.995\\
      GPUs & H100 $\times$ 1\\
      \bottomrule
    \end{tabular}
  }
  \caption{Hyperparameters used for training the prober.}
\label{tab:hyper-params}
\end{table}
We present the detailed hyperparameters for training the prober model. We use the first epoch as a warm-up epoch, and in the second epoch, we create a checkpoint using early stopping based on prober's classification accuracy.

\section{Prompt for Open-domain QA}
For all methods, we use the same 4-shot prompt as in Table~\ref{tab:prompt_retr}. When a retrieval call occurs, we include five retrieved passages in the Passages section. If directQA is required, we use the same prompt excluding the Passages section.
\begin{table}[]
\centering
\scalebox{0.85}{\begin{tabular}{l}
\hline
\begin{tabular}[c]{@{}p{1.0\linewidth}@{}}\\
\textbf{Query:} Who was the first President of the United States?\\
\textbf{Rationale:} The United States was formed after gaining independence from Britain. The first President would have been elected soon after the formation of the country. George Washington is commonly known as the leader of the American Revolution and the first President.\\
\textbf{Answer:} George Washington\\
\\
\textbf{Query:} Who wrote the play 'Romeo and Juliet'?\\
\textbf{Rationale:} “Romeo and Juliet” is a famous play, a tragedy involving two young lovers. The play is widely associated with English literature from the Renaissance period. William Shakespeare is the most renowned playwright of the English Renaissance and is known for his tragedies.\\
\textbf{Answer:} William Shakespeare\\
\\
\textbf{Query:} What is the main ingredient in traditional Italian pesto sauce?\\
\textbf{Rationale:} Pesto is a famous Italian sauce typically used in pasta dishes.\\ Traditional pesto originates from the Liguria region, specifically Genoa. The primary ingredients include basil, olive oil, garlic, pine nuts, and Parmesan cheese. The main flavoring herb that distinguishes pesto is basil.\\
\textbf{Answer:} Basil\\
\\
\textbf{Question:} Which writer was from England, Henry Roth or Robert Erskine Childers?\\
\textbf{Rationale:} Henry Roth was an American novelist, best known for his novel Call It Sleep (1934). He was born in Austria-Hungary (now Ukraine) and emigrated to the United States as a child. Robert Erskine Childers was an English-born writer and Irish nationalist. He was born in London, England, in 1870 and is best known for his novel The Riddle of the Sands (1903). Childers later became involved in Irish politics and was a prominent figure in the Irish independence movement.\\
\textbf{Answer:} Robert Erskine \\
\\
\textbf{Passages:} \\
\{\textit{Retrieved Passages}\}\\
\textbf{Question:}\\ 
\{\textit{New Query}\}\\
\textbf{Rationale:}\\

\end{tabular}  \\ \hline
\end{tabular}}
\caption{The 4-shot prompt used for Adaptive RAG methods.}
\label{tab:prompt_retr}
\end{table}

\section{Additional Analysis}
\label{appen:analysis}

\begin{table}[t]
\resizebox{\linewidth}{!}{\begin{tabular}{lccccccc}
\toprule
\multicolumn{1}{l||}{Threshold ($\theta$)}& HotpotQA& NQ & TriviaQA & MuSiQue &2Wiki \\ \midrule
\multicolumn{1}{l||}{-2} & 37.72 & 32.33 & 47.90 & 8.78 & 44.31 \\
\multicolumn{1}{l||}{-1} & 39.12 & 34.53 & 49.70 & 9.98 & 44.31 \\
\multicolumn{1}{l||}{0} & 39.12 & 35.53 & 50.50 & 9.98 & 43.71 \\
\multicolumn{1}{l||}{1} & 40.12 & 36.53 & 51.50 & 9.58 & 42.32\\
\multicolumn{1}{l||}{2} & 40.72 & 38.92 & 50.70 & 9.38 & 41.31 \\ \bottomrule
\end{tabular}}
\caption{Results among varying the classifying threshold for prober.}
\label{tab:threshold}
\end{table}

\subsection{Performance among Threshold of Prober}
\label{appen:threshold}
As mentioned in Section~\ref{sub:method_inference} and Algorithm~\ref{alg:probing}, we add a threshold value to $Logit[0]$, which determines whether to perform a retrieval call. A lower threshold reduces the likelihood of retrieval, while a higher threshold increases the retrieval frequency. In Table~\ref{tab:threshold}, we measure performance by varying the threshold from -2 to 2 in increments of 1. For HotpotQA, NaturalQA, and IIRC, increasing the threshold leads to better performance, whereas for TriviaQA, MuSiQue, and 2Wikimultihop datasets, performance either decreases or remains similar. This demonstrates that simply increasing the amount of retrieval does not improve QA performance.

\begin{table}[!ht]
\resizebox{\linewidth}{!}{\begin{tabular}{lccccccc}
\toprule
\multicolumn{1}{l||}{Probing Layer}& HotpotQA& NQ & TriviaQA & MuSiQue &2Wiki \\ \midrule
\multicolumn{1}{l||}{6} & 38.12 & 36.72 & 50.70 & 8.18 & 42.32  \\
\multicolumn{1}{l||}{6, 8} & 38.12 & 33.13 & 48.90 & 8.78 & 43.51  \\
\multicolumn{1}{l||}{6, 8, 10} & 38.12 & 32.34 & 48.70 & 8.98 & 44.31 \\
\multicolumn{1}{l||}{6, 8, 10, 12} & 38.92 & 33.73 & 49.30 & 9.38 & 44.31  \\
\multicolumn{1}{l||}{6, 8, 10, 12, 14} & 39.12 & 34.53 & 49.70 & 9.98 & 44.31  \\ \bottomrule
\end{tabular}}
\caption{Accuracy among the prober location within the layers.}
\label{tab:ablation}
\end{table}
\subsection{Impact of Varying the Number of Prober Layers on QA Performance}
\label{appen:ablation}
We use the prober’s logits to decide whether to perform retrieval. Summing the logits from each layer’s prober provides a soft voting effect and enhances model performance, as empirically shown in Table~\ref{tab:ablation}. We measure performance while removing one probing layer at a time. We observe that having more probers increases performance across most datasets except for NaturalQA and TriviaQA. Using probers up to layer 14 improves the average performance across most datasets by 2.0 percentage points compared to using a single prober at layer 6.

\subsection{Depth-wise Comparison between Probing-RAG and DRAGIN}
\label{appen:em}
\begin{figure*}[htbp]
\centerline{\includegraphics[scale=0.335]{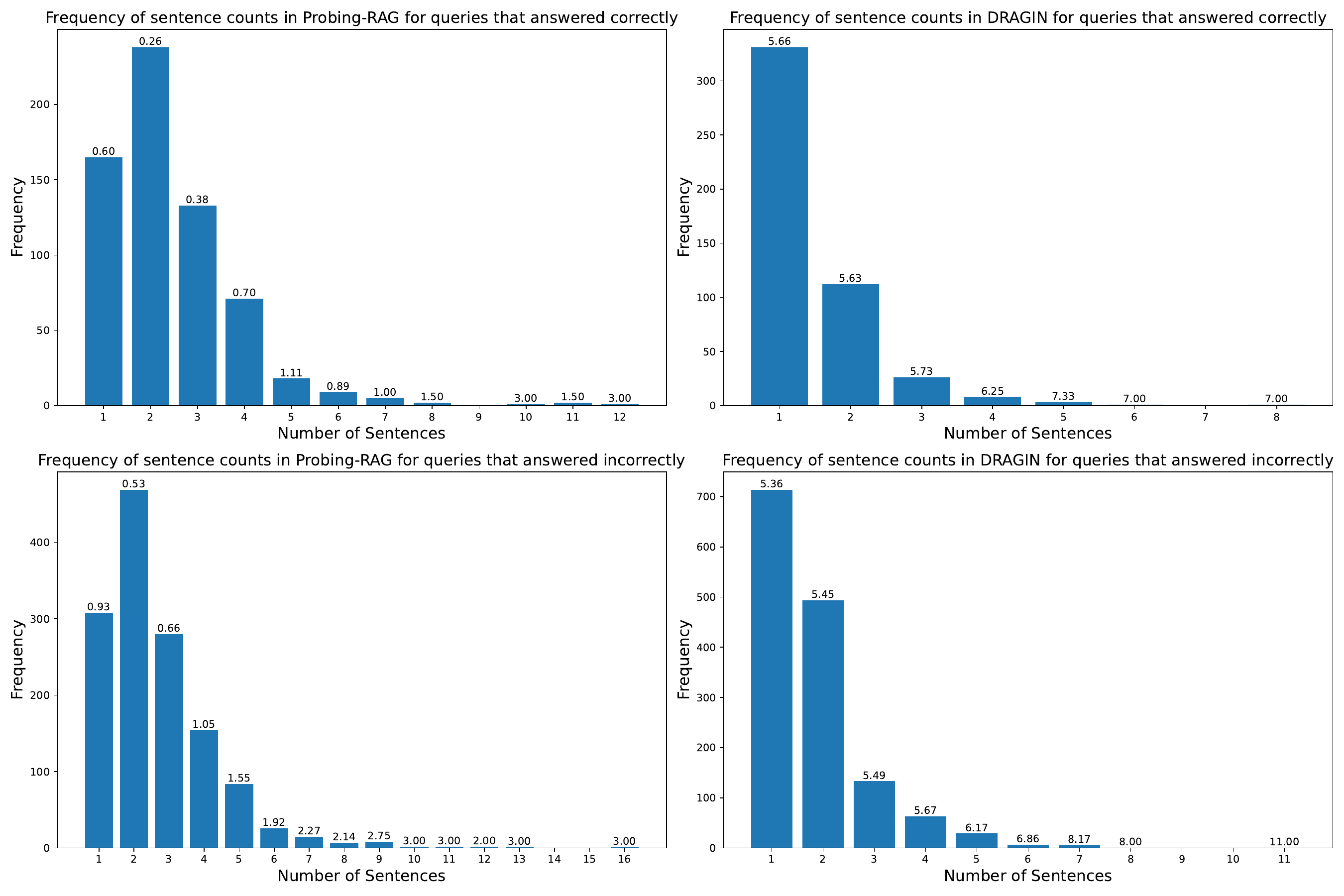}}
\caption{The numbers above the bar plots represent the average number of retrieval calls. The top-left section shows the number of rationale sentences when Probing-RAG answered correctly, and the bottom-left shows when it answered incorrectly. The top-right indicates the number of rationale sentences when DRAGIN answered correctly, while the bottom-right indicates when it answered incorrectly.}
\label{fig:number_of_sentence}
\end{figure*}
In Figure~\ref{fig:number_of_sentence}, the numbers above the bar plots represent the average number of retrieval calls. The upper left shows the number of rationale sentences when Probing-RAG answers correctly, and the lower left shows the number when Probing-RAG answers incorrectly. The upper-right displays the number of rationale sentences when DRAGIN answers correctly, and the lower-right shows the number when DRAGIN answers incorrectly. As shown in Table~\ref{tab:main}, DRAGIN has lower accuracy but higher exact match performance. To explain this, we compare the outputs of DRAGIN and Probing-RAG in the case study presented in Table~\ref{tab:case}. For a more substantial explanation, we analyze statistically the answers generated by DRAGIN and Probing-RAG to understand the reason behind the high exact match performance. The top-left and top-right of Figure~\ref{fig:number_of_sentence} shows the number of sentences in the rationale sections for Probing-RAG and DRAGIN, respectively. Probing-RAG makes 0.6 and 0.26 retrieval calls when generating one and two sentences, respectively. In contrast, DRAGIN makes 5.66 and 5.63 retrieval calls. Making more than five retrieval calls to generate a single sentence enables accurate answers based on external knowledge. As shown in Table~\ref{tab:case}, DRAGIN directly references external knowledge to produce its responses. DRAGIN ignores questions that the language model could answer using internal knowledge and heavily relies on external information. This reliance leads to decreased accuracy when external knowledge is inaccessible. In contrast, Probing-RAG makes fewer than one retrieval call, generating correct answers without depending on external knowledge for questions answerable through internal knowledge. This approach decreases exact match performance but increases accuracy when the model uses its internal knowledge to generate answers. We also observe similar sentence lengths and numbers of retrieval calls even when incorrect answers are generated. Additionally, Adaptive-RAG methods typically perform more retrievals for complex questions. However, DRAGIN makes fewer retrieval calls when it produces incorrect answers than when it answers correctly. In contrast, Probing-RAG makes more retrieval calls in situations where it generates incorrect answers. Through the statistical analysis in Figure~\ref{fig:number_of_sentence}, we explain why DRAGIN exhibits high Exact Match performance but low accuracy. DRAGIN relies heavily on external knowledge by making multiple inefficient retrieval calls. In contrast, Probing-RAG significantly improves accuracy by efficiently utilizing retrieval and leveraging internal knowledge.

\subsection{Additional Case Study}
\label{appen:case}
To enhance the case study section, we add two examples from each of the four datasets used in our evaluation: NaturalQA, HotpotQA, TriviaQA, and MuSiQue. We provide the rationale and answers generated by the Probing RAG, Adaptive RAG, DRAGIN, and FLARE methods, along with their accuracy and number of retrievals. Here, an accuracy of 1 indicates a correct answer, and 0 indicates an incorrect one.

\begin{table*}[t]
\centering
\scalebox{0.85}{\begin{tabular}{ll}
\hline
\multicolumn{2}{c}{\textbf{NaturalQA Example 1}}\\
\multicolumn{2}{c}{\textbf{Question:} which apostle had a thorn in his side}\\
\multicolumn{2}{c}{\textbf{Answer:} Paul}\\
\midrule
\textbf{Probing-RAG}        & \begin{tabular}[c]{@{}p{1.0\linewidth}@{}}
\textbf{Rationale:} The apostle Paul had a thorn in his side. This is a reference to 2 Corinthians 12:7-10.\\
\textbf{Answer:} Paul \\
\textbf{Accuracy:}  1 \\
\textbf{Number of Total Retrieval Call:}  0 \\

\end{tabular}  \\ \midrule
\textbf{Adaptive-RAG}        & \begin{tabular}[c]{@{}p{1.0\linewidth}@{}}
\textbf{Rationale:} Paul the Apostle was a prominent figure in the early Christian church. He was a follower of Jesus Christ and was known for his missionary work. He was also known for his writings in the New Testament, particularly his letters to the early Christian communities. Paul was a prolific writer and is considered one of the most influential figures in the history of Christianity. He is known for his teachings on topics such as salvation, grace, and the Christian life.\\
\textbf{Answer:} Paul the Apostle \\
\textbf{Accuracy:}  1 \\
\textbf{Number of Total Retrieval Call:}  1 \\

\end{tabular}  \\ \midrule
\textbf{DRAGIN}        & \begin{tabular}[c]{@{}p{1.0\linewidth}@{}}
\textbf{Rationale:} Paul the Apostle had a thorn in his side. This is stated in 2 Corinthians 12:7-9. \\
\textbf{Answer:} Paul the Apostle \\
\textbf{Accuracy:}  1 \\
\textbf{Number of Total Retrieval Call:}  6 \\

\end{tabular}  \\ \midrule
\textbf{FLARE}        & \begin{tabular}[c]{@{}p{1.0\linewidth}@{}}
\textbf{Rationale:} Paul the Apostle was a prominent figure in the early Christian church. He was a follower of Jesus Christ and was known for his missionary work. He was also known for his writings in the New Testament, particularly his letters to the early Christian communities. Paul was a prolific writer and is considered one of the most influential figures in the history of Christianity. He is known for his teachings on topics such as salvation, grace, and the Christian life. \\
\textbf{Answer:} Paul \\
\textbf{Accuracy:}  1 \\
\textbf{Number of Total Retrieval Call:}  5 \\
\end{tabular}  \\ 
\midrule
\midrule
\multicolumn{2}{c}{\textbf{NaturalQA Example 2}}\\
\multicolumn{2}{c}{\textbf{Question:} who sang the original version of true colors}\\
\multicolumn{2}{c}{\textbf{Answer:} Cyndi Lauper }\\
\midrule
\textbf{Probing-RAG}        & \begin{tabular}[c]{@{}p{1.0\linewidth}@{}}
\textbf{Rationale:} The original version of "True Colors" was sung by Cyndi Lauper. The song was released in 1986 and became a hit, reaching number one on the Billboard Hot 100 chart.\\
\textbf{Answer:} Cyndi Lauper \\
\textbf{Accuracy:}  1 \\
\textbf{Number of Total Retrieval Call:}  0 \\

\end{tabular}  \\ \midrule
\textbf{Adaptive-RAG}        & \begin{tabular}[c]{@{}p{1.0\linewidth}@{}}
\textbf{Rationale:} True Colors is a song by Russian-German electronic music producer Zedd. The original version of the song featured uncredited vocals by Tim James and was included on Zedd's second studio album, "True Colors" (2015). The official single is a new version of the track with vocals provided by the American singer Kesha and was released as the album's fourth single on April 29, 2016. They performed the song live at Coachella 2016. The song appeared in "FIFA 17". "True Colors" was Kesha's first music release in over three years, as she has been locked.\\
\textbf{Answer:} Kesha \\
\textbf{Accuracy:}  0 \\
\textbf{Number of Total Retrieval Call:}  1 \\

\end{tabular}  \\ \midrule
\textbf{DRAGIN}        & \begin{tabular}[c]{@{}p{1.0\linewidth}@{}}
\textbf{Rationale:} The original version of "True Colors" was sung by Cyndi Lauper.  \\
\textbf{Answer:} Cyndi Lauper  \\
\textbf{Accuracy:}  1 \\
\textbf{Number of Total Retrieval Call:}  7 \\

\end{tabular}  \\ \midrule
\textbf{FLARE}        & \begin{tabular}[c]{@{}p{1.0\linewidth}@{}}
\textbf{Rationale:} The song was released on 29 October 2010. \\
\textbf{Answer:} Kesha \\
\textbf{Accuracy:}  0 \\
\textbf{Number of Total Retrieval Call Call:} 1 \\
\end{tabular}  \\
\midrule
\end{tabular}}
\caption{Output examples for the NaturalQA dataset.}
\label{tab:nq_example}
\end{table*}
\begin{table*}[t]
\centering
\scalebox{0.8}{\begin{tabular}{ll}
\hline
\multicolumn{2}{c}{\textbf{HotpotQA Example 1}}\\
\multicolumn{2}{c}{\textbf{Question:} Which dog's ancestors include Gordon and Irish Setters: the Manchester Terrier or the Scotch Collie?}\\
\multicolumn{2}{c}{\textbf{Answer:} Scotch Collie }\\
\midrule
\textbf{Probing-RAG}        & \begin{tabular}[c]{@{}p{1.0\linewidth}@{}}
\textbf{Rationale:} The Manchester Terrier is a breed of dog that originated in England. It is a small, sturdy dog with a short, smooth coat. The Scotch Collie is a breed of dog that originated in Scotland. It is a medium-sized dog with a long, straight coat.\\
\textbf{Answer:} Scotch Collie \\
\textbf{Accuracy:}  1 \\
\textbf{Number of Total Retrieval Call:}  0 \\

\end{tabular}  \\ \midrule
\textbf{Adaptive-RAG}        & \begin{tabular}[c]{@{}p{1.0\linewidth}@{}}
\textbf{Rationale:} The Manchester Terrier is a breed of dog that originated in England. It is a small, sturdy dog with a short, smooth coat. The Manchester Terrier is known for its intelligence, loyalty, and playful nature. The Scotch Collie is a breed of dog that originated in Scotland. It is a medium-sized dog with a long, smooth coat. The Scotch Collie is known for its intelligence, loyalty, and playful nature.\\
\textbf{Answer:} Manchester Terrier  \\
\textbf{Accuracy:}  0 \\
\textbf{Number of Total Retrieval Call:}  2 \\

\end{tabular}  \\ \midrule
\textbf{DRAGIN}        & \begin{tabular}[c]{@{}p{1.0\linewidth}@{}}
\textbf{Rationale:} The Manchester Terrier is a breed of dog of the smooth-haired terrier type. The Scotch Collie is a breed of dog, categorized as a terrier.  \\
\textbf{Answer:} Scotch Collie  \\
\textbf{Accuracy:}  1 \\
\textbf{Number of Total Retrieval Call:}  5 \\

\end{tabular}  \\ \midrule
\textbf{FLARE}        & \begin{tabular}[c]{@{}p{1.0\linewidth}@{}}
\textbf{Rationale:} The Manchester Terrier is a breed of dog of the smooth-haired terrier type. The Irish Red and White Setter is a medium-sized dog with a long, silky coat. The Irish Setter is a medium-sized dog with a long, silky coat. The Scotch Collie is a landrace breed of dog which originated from the highland regions of Scotland. The Manchester Terrier is a breed of dog of the smooth-haired terrier type. \\
\textbf{Answer:} Scotch Collie \\
\textbf{Accuracy:}  1 \\
\textbf{Number of Total Retrieval Call:}  5 \\
\end{tabular}  \\ 
\midrule
\midrule
\multicolumn{2}{c}{\textbf{HotpotQA Example 2}}\\
\multicolumn{2}{c}{\textbf{Question:} Alfred Balk served as the secretary of the Committee on the Employment of}\\
\multicolumn{2}{c}{Minority Groups in the News Media under which United States Vice}\\
\multicolumn{2}{c}{\textbf{Answer:} Nelson Rockefeller  }\\
\midrule
\textbf{Probing-RAG}        & \begin{tabular}[c]{@{}p{1.0\linewidth}@{}}
\textbf{Rationale:} The Committee on the Employment of Minority Groups in the News Media was established by the United States Vice President Nelson Rockefeller in 1972. The committee was established to study the employment of minorities in the news media. The committee was chaired by Alfred Balk.\\
\textbf{Answer:} Nelson Rockefeller  \\
\textbf{Accuracy:}  1 \\
\textbf{Number of Total Retrieval Call:}  3 \\

\end{tabular}  \\ \midrule
\textbf{Adaptive-RAG}        & \begin{tabular}[c]{@{}p{1.0\linewidth}@{}}
\textbf{Rationale:} Alfred Balk was a United States Vice President. He served as the Secretary of the Committee on the Employment of Minority Groups in the News Media under President Lyndon B. Johnson.
Answer: Lyndon B. Johnson.\\
\textbf{Answer:} Lyndon B. Johnson  \\
\textbf{Accuracy:}  0 \\
\textbf{Number of Total Retrieval Call:}  2 \\

\end{tabular}  \\ \midrule
\textbf{DRAGIN}        & \begin{tabular}[c]{@{}p{1.0\linewidth}@{}}
\textbf{Rationale:} The National News Council was established by the Twentieth Century Fund in 1965. The committee was established to study the employment of minorities in the news media. Alfred Balk was the secretary of the committee. 
Answer: Lyndon B. Johnson.  \\
\textbf{Answer:} Lyndon B. Johnson  \\
\textbf{Accuracy:}  0 \\
\textbf{Number of Total Retrieval Call:}  8 \\

\end{tabular}  \\ \midrule
\textbf{FLARE}        & \begin{tabular}[c]{@{}p{1.0\linewidth}@{}}
\textbf{Rationale:} The Vice President of the United States is the second-highest-ranking official in the United States federal government, after the President of the United States. The Vice President is the President of the Senate and is the presiding officer of the Senate. The Vice President is the President of the United States in the event of the death, resignation, or removal in office of the President. \\
\textbf{Answer:} Lyndon B. Johnson  \\
\textbf{Accuracy:}  0 \\
\textbf{Number of Total Retrieval Call Call:} 3 \\
\end{tabular}  \\
\midrule
\end{tabular}}
\caption{Output examples for the HotpotQA dataset.}
\label{tab:hotpot_example}
\end{table*}
\begin{table*}[t]
\centering
\scalebox{0.85}{\begin{tabular}{ll}
\hline
\multicolumn{2}{c}{\textbf{TriviaQA Example 1}}\\
\multicolumn{2}{c}{\textbf{Question:} Which ship was sunk by a German submarine in 1915 with a loss of 1400 lives}\\
\multicolumn{2}{c}{including 128 Americans?}\\
\multicolumn{2}{c}{\textbf{Answer:} [Roman Lusitania, Luso-, Lusitânia, Lusitania, Hispania Lusitania,}\\
\multicolumn{2}{c}{Lusitania (ancient region), Roman province Lusitania, Lusitania (Roman province)]}\\
\midrule
\textbf{Probing-RAG}        & \begin{tabular}[c]{@{}p{1.0\linewidth}@{}}
\textbf{Rationale:} The Lusitania was a British passenger liner that was sunk by a German submarine in 1915. The Lusitania was carrying 128 Americans, including 128 Americans, and 128 British passengers. The Lusitania was sunk by a German submarine, and the loss of life was a major factor in the United States' decision to enter World War I.\\
\textbf{Answer:} Lusitania \\
\textbf{Accuracy:}  1 \\
\textbf{Number of Total Retrieval Call:}  0 \\

\end{tabular}  \\ \midrule
\textbf{Adaptive-RAG}        & \begin{tabular}[c]{@{}p{1.0\linewidth}@{}}
\textbf{Rationale:} The Lusitania was a British passenger liner that was sunk by a German submarine in 1915. The Lusitania was carrying 1,959 passengers and crew, including 128 Americans. The Lusitania was torpedoed by the German submarine U-20, killing 1,198 people, including 128 Americans. The Lusitania was the first major loss of life in the First World War.\\
\textbf{Answer:} Lusitania\\
\textbf{Accuracy:}  1 \\
\textbf{Number of Total Retrieval Call:}  1 \\

\end{tabular}  \\ \midrule
\textbf{DRAGIN}        & \begin{tabular}[c]{@{}p{1.0\linewidth}@{}}
\textbf{Rationale:} The RMS Lusitania was a British passenger liner that was sunk by a German submarine in 1915. The Lusitania was the largest passenger ship in the world at the time of its sinking. The Lusitania was carrying 1,959 passengers and crew when it was sunk. The Lusitania was the largest passenger ship in the world at the time of its sinking.  \\
\textbf{Answer:} Lusitania\\
\textbf{Accuracy:}  1 \\
\textbf{Number of Total Retrieval Call:}  5 \\

\end{tabular}  \\ \midrule
\textbf{FLARE}        & \begin{tabular}[c]{@{}p{1.0\linewidth}@{}}
\textbf{Rationale:} The Lusitania was a British passenger liner that was sunk by a German submarine in 1915. The sinking of the "Lusitania" was the event that decisively swung American opinion against Germany. The sinking of the "Lusitania" was the event that decisively swung American opinion against Germany.\\
\textbf{Answer:} Lusitania \\
\textbf{Accuracy:}  1 \\
\textbf{Number of Total Retrieval Call:}  3 \\
\end{tabular}  \\ 
\midrule
\midrule
\multicolumn{2}{c}{\textbf{TriviaQA Example 2}}\\
\multicolumn{2}{c}{\textbf{Question:} Balice international airport is in which country?}\\
\multicolumn{2}{c}{\textbf{Answer:} [Etymology of Poland, Polish state, Poleand, Rzeczpospolita Polska, Polland, ISO 3166-1:PL, Polskor,}\\
\multicolumn{2}{c}{Republic of Poland, POLAND, Fourth Poland, Country- Poland, Poleland, Polska, Poland, Poland., Pols Land,} \\
\multicolumn{2}{c}{Polnd, Po land, Lenkija, Bastarnia, Pologne, PolanD, Polija, Pole land, Architecture of Poland, Ploand]}\\
\midrule
\textbf{Probing-RAG}        & \begin{tabular}[c]{@{}p{1.0\linewidth}@{}}
\textbf{Rationale:} Balice international airport is in the country of Croatia.\\
\textbf{Answer:} Croatia \\
\textbf{Accuracy:}  0 \\
\textbf{Number of Total Retrieval Call:}  0 \\

\end{tabular}  \\ \midrule
\textbf{Adaptive-RAG}        & \begin{tabular}[c]{@{}p{1.0\linewidth}@{}}
\textbf{Rationale:} The airport is located in the village of Balice, west of the city centre, in southern Poland.\\
\textbf{Answer:} Poland \\
\textbf{Accuracy:}  1 \\
\textbf{Number of Retrieval Call:}  1 \\

\end{tabular}  \\ \midrule
\textbf{DRAGIN}        & \begin{tabular}[c]{@{}p{1.0\linewidth}@{}}
\textbf{Rationale:} Balice international airport is in Poland.   \\
\textbf{Answer:} Poland  \\
\textbf{Accuracy:}  1 \\
\textbf{Number of Total Retrieval Call:}  7 \\

\end{tabular}  \\ \midrule
\textbf{FLARE}        & \begin{tabular}[c]{@{}p{1.0\linewidth}@{}}
\textbf{Rationale:} The airport is located in the village of Balice, west of the city centre, in southern Poland. \\
\textbf{Answer:} Poland \\
\textbf{Accuracy:}  1 \\
\textbf{Number of Total Retrieval Call Call:} 1 \\
\end{tabular}  \\
\midrule
\end{tabular}}
\caption{Output examples for the TriviaQA dataset.}
\label{tab:trivia_example}
\end{table*}
\begin{table*}[t]
\centering
\scalebox{0.85}{\begin{tabular}{ll}
\hline
\multicolumn{2}{c}{\textbf{MuSiQue Example 1}}\\
\multicolumn{2}{c}{\textbf{Question:} Who wrote "Turn Me On" by performer of "Happy Pills"?}\\
\multicolumn{2}{c}{\textbf{Answer:} John D. Loudermilk}\\
\midrule
\textbf{Probing-RAG}        & \begin{tabular}[c]{@{}p{1.0\linewidth}@{}}
\textbf{Rationale:} "Turn Me On" is a song by Norah Jones from her album First Sessions / Come Away with Me. The song was written by John D. Loudermilk.\\
\textbf{Answer:} John D. Loudermilk \\
\textbf{Accuracy:}  1 \\
\textbf{Number of Total Retrieval Call:}  1 \\

\end{tabular}  \\ \midrule
\textbf{Adaptive-RAG}        & \begin{tabular}[c]{@{}p{1.0\linewidth}@{}}
\textbf{Rationale:}  "Turn Me On" is a song by Norah Jones from her album First Sessions / Come Away with Me. The song was written by John D. Loudermilk.\\
\textbf{Answer:} John D. Loudermilk \\
\textbf{Accuracy:}  1 \\
\textbf{Number of Total Retrieval Call:}  1 \\

\end{tabular}  \\ \midrule
\textbf{DRAGIN}        & \begin{tabular}[c]{@{}p{1.0\linewidth}@{}}
\textbf{Rationale:} The performer of "Happy Pills" is the American singer-songwriter and actress, Cher. She is best known for her work in the 1960s and 1970s, and for her work in the 1980s and 1990s. She has sold over 100 million records worldwide, making her one of the best-selling music artists of all time.  \\
\textbf{Answer:} Cher \\
\textbf{Accuracy:}  0 \\
\textbf{Number of Total Retrieval Call:}  4 \\

\end{tabular}  \\ \midrule
\textbf{FLARE}        & \begin{tabular}[c]{@{}p{1.0\linewidth}@{}}
\textbf{Rationale:} The Beatles' first album, Please Please Me, was released in the United Kingdom on 22 March 1963. The Beatles' White Album was released in the United States on 15 April 1963. The first episode of the series aired on 17 December 1999. \\
\textbf{Answer:} The Beatles \\
\textbf{Accuracy:}  0 \\
\textbf{Number of Total Retrieval Call:}  3 \\
\end{tabular}  \\ 
\midrule
\midrule
\multicolumn{2}{c}{\textbf{MuSiQue Example 2}}\\
\multicolumn{2}{c}{\textbf{Question:} What is the date of birth of the person who was part of Ratata?}\\
\multicolumn{2}{c}{\textbf{Answer:} 11 September 1962 }\\
\midrule
\textbf{Probing-RAG}        & \begin{tabular}[c]{@{}p{1.0\linewidth}@{}}
\textbf{Rationale:} Ratata was a pop group formed in 1980. The group was composed of Mauro Scocco and Johan Ekelund. The group was active from 1980 to 1989.\\
\textbf{Answer:} 11 September 1962 \\
\textbf{Accuracy:}  1 \\
\textbf{Number of Total Retrieval Call:}  1 \\

\end{tabular}  \\ \midrule
\textbf{Adaptive-RAG}        & \begin{tabular}[c]{@{}p{1.0\linewidth}@{}}
\textbf{Rationale:} Ratata was a Polish resistance organization that was active during World War II. The organization was founded in 1942 by Augustyn Träger (25 August 1896 – 22 April 1957), codenames "Sęk" (Knot) and "Tragarz" (Porter), and Roman Träger (1922–1944). The organization was active in the Polish resistance movement in the occupied territories of Poland.\\
\textbf{Answer:} 25 August 1896  \\
\textbf{Accuracy:}  0 \\
\textbf{Number of Total Retrieval Call:}  2 \\

\end{tabular}  \\ \midrule
\textbf{DRAGIN}        & \begin{tabular}[c]{@{}p{1.0\linewidth}@{}}
\textbf{Rationale:} The answer is Peter Fliesteden. \\
\textbf{Answer:} Peter Fliesteden   \\
\textbf{Accuracy:}  0 \\
\textbf{Number of Total Retrieval Call:}  6 \\

\end{tabular}  \\ \midrule
\textbf{FLARE}        & \begin{tabular}[c]{@{}p{1.0\linewidth}@{}}
\textbf{Rationale:} Ratata was a pop group formed in 1980. The group was composed of Mauro Scocco and Johan Ekelund. The group was active from 1980 to 1989. \\
\textbf{Answer:} 11 September 1962 \\
\textbf{Accuracy:}  1 \\
\textbf{Number of Total Retrieval Call Call:} 3 \\
\end{tabular}  \\
\midrule
\end{tabular}}
\caption{Output examples for the MuSiQue dataset.}
\label{tab:musique_example}
\end{table*}
\newpage
\clearpage

\end{document}